\documentclass[runningheads]{llncs}
\usepackage{graphicx}
\usepackage{amsmath}
\usepackage{booktabs}
\usepackage{colortbl}
\usepackage{amssymb}
\usepackage{subcaption}
\usepackage{multirow}
\usepackage{marvosym}

\begin{document}
\title{SLENet: A Guidance-Enhanced Network for Underwater Camouflaged Object Detection
% \thanks{Supported by organization x.}
}
\titlerunning{SLENet for Underwater Camouflaged Object Detection}
\author{
    Xinxin Huang\inst{1} \and 
    Han Sun\inst{1}\textsuperscript{(\Letter)} \and 
    Ningzhong Liu\inst{1} \and 
    Huiyu Zhou\inst{2} \and 
    Yinan Yao\inst{1}
}
% \author{Anonymous Authors}
% \institute{}

%
\authorrunning{X. Huang et al.}
% \authorrunning{Anonymous Authors}
\institute{
    % (inst{1})
    Nanjing University of Aeronautics and Astronautics, Nanjing, China \\
    % \email{xxhuang2002@163.com, sunhan@nuaa.edu.cn, , 18858357106@163.com} % 请替换为真实的邮箱前缀
    \email{xxhnuaa@163.com,~sunhan@nuaa.edu.cn,~lnz\_nuaa@163.com,~yyn\_nuaa@163.com}
    \and
    % (inst{2})
    University of Leicester, Leicester, UK \\
    \email{hz143@leicester.ac.uk}
}
\maketitle              % typeset the header of the contribution
\begin{abstract} Underwater Camouflaged Object Detection (UCOD) aims to identify objects that blend seamlessly into underwater environments. This task is critically important to marine ecology. However, it remains largely underexplored and accurate identification is severely hindered by optical distortions, water turbidity, and the complex traits of marine organisms. To address these challenges, we introduce the UCOD task and present DeepCamo, a benchmark dataset designed for this domain. We also propose Semantic Localization and Enhancement Network (SLENet), a novel framework for UCOD. We first benchmark state-of-the-art COD models on DeepCamo to reveal key issues, upon which SLENet is built. In particular, we incorporate Gamma-Asymmetric Enhancement (GAE) module and a Localization Guidance Branch (LGB) to enhance multi-scale feature representation while generating a location map enriched with global semantic information. This map guides the Multi-Scale Supervised Decoder (MSSD) to produce more accurate predictions. Experiments on our DeepCamo dataset and three benchmark COD datasets confirm SLENet's superior performance over SOTA methods, and underscore its high generality for the broader COD task.

\end{abstract}
\section{Introduction}
\begin{figure}
\includegraphics[width=\textwidth]{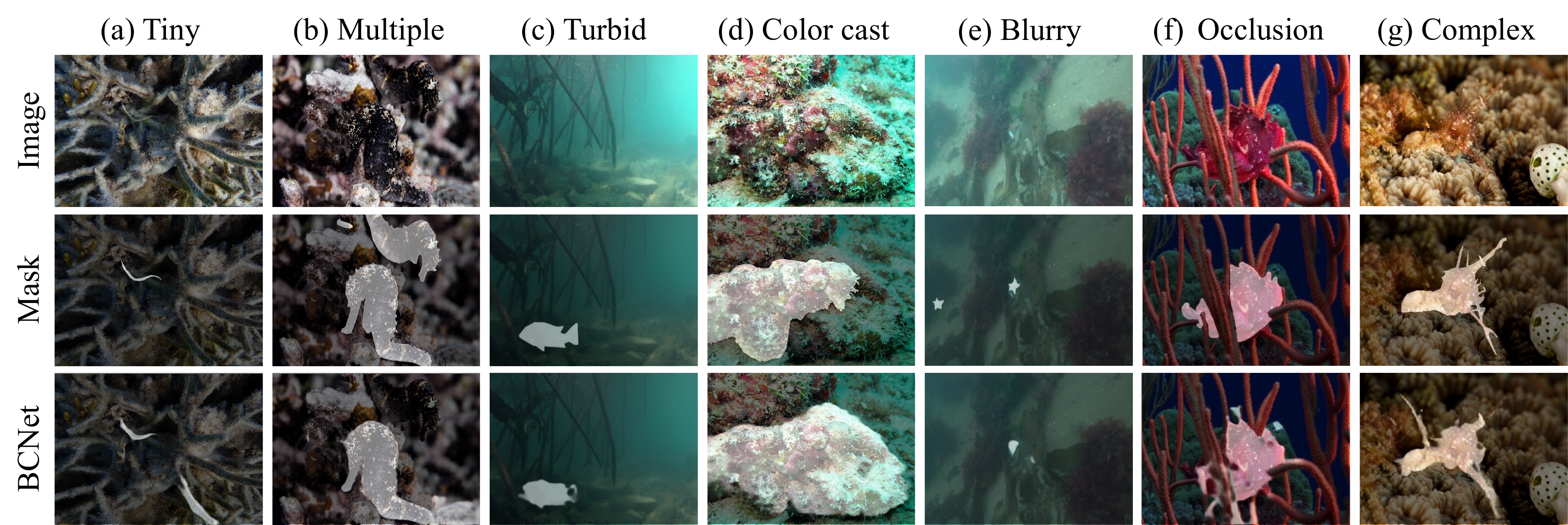}
\caption{Our dataset features challenging scenarios, including low-quality images (c,d,e), complex object attributes (a,b,g), and distracting backgrounds (f), where general COD methods like BCNet often suffer from missed detections or produce unclear boundaries.} \label{first}
\end{figure}

Detecting camouflaged objects in underwater environments presents a unique and compounded set of challenges. Beyond the inherent difficulty of identifying objects that mimic their surroundings' color and texture, this task is further complicated by two factors: the complex characteristics of marine organisms (e.g., small size, large quantities, intricate boundaries) and severe image degradation from optical distortion and color cast. UCOD is a critical yet largely unaddressed issue for marine ecological monitoring and biodiversity conservation, with no systematic research to date in this domain. Consequently, existing general-purpose COD methods~\cite{fan2020camouflaged,mei2021PFnet,hu2023hitnet,2023FEDER,2023fsnet,xiao2023HFEM,fan2021SINet-v2,xiong2024sam2UNet} often struggle to handle these unique underwater challenges, as shown in Fig.~\ref{first}.

\noindent Deep learning excels at COD with powerful feature extraction, surpassing traditional methods reliant on handcrafted features. Modeling complex underwater scenes requires capturing comprehensive context and long-range dependencies, yet common feature enhancement techniques often sacrifice fine details to achieve this. Models like the Segment Anything Model (SAM1)~\cite{kirillov2023SAM1} and SAM2~\cite{ravi2024sam2} have improved segmentation performance, but their simple decoders often struggle in downstream tasks, frequently segmenting irrelevant regions. Consequently, existing COD methods often fail to precisely locate and segment small or multiple underwater objects. Furthermore, existing COD datasets~\cite{fan2020camouflaged,le2019CAMO,2018Chameleon} have limited diversity in underwater scenes and often overlook multi-object scenarios, which constrains model generalization and performance on UCOD tasks.

To address the above challenges, we construct DeepCamo, a dataset focused on underwater camouflaged objects in complex marine environments. Building upon the powerful encoding capability of SAM2, we propose Semantic Localization and Enhancement Network (SLENet) for the UCOD task. We use the SAM2 encoder as a backbone, freezing its pretrained weights and introducing Adapters for efficient fine-tuning. To enrich multi-scale features while preserving fine details, we introduce a cascaded Gamma-Asymmetric Enhancement (GAE) Module to enhance features at each hierarchical level. Inspired by biological hunting mechanisms, a Localization Guidance Branch (LGB) fuses multi-scale features bottom-up to generate a coarse localization map. This map then injects global localization cues into the Multi-Scale Supervised Decoder (MSSD), enabling accurate object localization. Experimental results demonstrate that SLENet significantly outperforms existing SOTA methods on the UCOD task. In summary, the main contributions of this paper are as follows:
\begin{itemize}
    \item We conduct a systematic exploration of the UCOD task and introduce DeepCamo, a new UCOD benchmark containing 2,493 images of 16 marine species, covering diverse illumination conditions and multi-object scenarios.
    \item We propose the SLENet as a novel UCOD framework that effectively adapts the SAM2 encoder. It integrates the GAE to enrich features and an LGB to generate a global localization map, which in turn provides precise spatial cues to the MSSD.
    \item We validate each proposed module through extensive experiments, demonstrating that our SLENet consistently outperforms SOTA methods on our DeepCamo dataset and three other public COD benchmarks.
\end{itemize}
\section{Related work}
\subsection{Camouflaged object detection}
Camouflaged object detection (COD) aims to segment objects that resemble their background in color, texture, and shape. Traditional methods using handcrafted low-level features perform adequately in simple scenes but struggle in complex environments. The availability of large-scale COD datasets~\cite{fan2020camouflaged} has enabled deep learning approaches to become mainstream. Fan et al.~\cite{fan2020camouflaged} introduced SINet, a bio-inspired CNN framework using a two-stage search-and-identify strategy. PFNet~\cite{mei2021PFnet} employs high-level features for object localization and utilizes focus modules to reduce distractions. However, CNNs' limited feature extraction hinders discerning subtle foreground-background differences, especially in underwater scenes with severe color bias.

Vision Transformer(ViT)~\cite{2020ViT}, capable of modeling long-range dependencies and capturing global context, is a leading COD solution. FSPNet~\cite{huang2023fspnet} employs a pyramid structure to progressively compress adjacent transformer features, accumulating informative representations. HitNet~\cite{hu2023hitnet} introduces an iterative feedback mechanism where high-resolution features refine low-resolution representations, alleviating detail degradation. Nevertheless, Transformers require large-scale data to perform well, while underwater camouflaged images are scarce. To address this, we fine-tuning pre-trained visual foundation models to extract global features while preserving details, and incorporate localization-guided strategies to achieve accurate segmentation of underwater camouflaged objects.

\subsection{Segment Anything Model for COD }
Recently, SAM1~\cite{kirillov2023SAM1} demonstrated strong zero-shot image segmentation capabilities. Its successor, SAM2~\cite{ravi2024sam2}, was trained on an expanded dataset and incorporates various enhancements. However, both models, trained on natural images, struggle with subtle foreground-background distinctions in complex underwater camouflaged scenes. To address this, many studies employ Adapters~\cite{houlsby2019adapter1,qiu2023lAdapter2} for parameter-efficient fine-tuning, offering domain-specific guidance. Yet, the native decoders of SAM1 and SAM2 remain too simplistic for complex tasks like COD. To improve performance, SAM2-UNet~\cite{xiong2024sam2UNet} integrates SAM2's encoder into a U-Net architecture for efficient camouflage detection. Nonetheless, UCOD remains underexplored. Consequently, we propose a UCOD-oriented framework retaining SAM2's Adapter-equipped hierarchical encoder, enabling effective capture of UCOD-specific prior knowledge.

\subsection{Multi-scale Feature Enhancement }
Multi-scale context is crucial in object detection, as it enhances feature representation. Hu et al.~\cite{hu2021dense} thus proposed parallel pooling at various resolutions for multi-scale features. PoolNet~\cite{liu2019poolnet} used varied downsampling rate pooling layers for multi-scale local context. While low-resolution features offer multi-scale information, they often cause detail loss. Chen et al.~\cite{chen2017deeplabASPP} addressed this with dilated convolutions, expanding receptive fields without extra cost or lower resolution, and developed ASPP to capture multi-scale context via varied dilation rate convolutions. Motivated by this, many COD methods adopt similar strategies. For instance, RFB~\cite{wu2019RFB} employed parallel dilated convolutions to expand receptive fields, mimicking human vision. However, large-dilation convolutions tend to lose fine details, challenging detection of underwater camouflaged objects with drastic scale changes. HFEM~\cite{xiao2023HFEM} captures multi-scale features via a serial multi-branch structure, avoiding large dilation rates to preserve details. However, its feature extraction and efficiency remain limited. To address this, our proposed GAE combines asymmetric and rate-2 dilated convolutions to enhance directional perception and achieve efficient multi-scale feature extraction.

\section{Datasets and Benchmark Studies}
\subsection{Data Collection}
Underwater images from common camouflage datasets often fail to capture the unique challenges of marine environments, such as color distortion, optical aberrations, and water flow effects. Moreover, these datasets lack multi-object scenarios, as the images are predominantly single-object. This data bias is a critical issue, as real marine organisms often appear in groups, which limits a model's generalization ability to complex scenes.

To this end, we introduce DeepCamo, a UCOD dataset aptly named to reflect both the physical underwater depth and the depth of camouflage. It is curated by selecting camouflaged images from underwater datasets such as MAS3K~\cite{li2020mas3k}, RMAS~\cite{fu2023Rmas}, and UFO120~\cite{UFO120}, and merging them with relevant underwater scenes from the camouflage datasets CAMO~\cite{le2019CAMO}, COD10K~\cite{fan2020camouflaged}, and CHAMELEON~\cite{2018Chameleon}. 

DeepCamo’s design follows several key principles. It emphasizes seamless camouflage, making objects difficult to discern from their background, and incorporates challenging underwater environments with issues like color distortion, blur, and occlusion. Furthermore, for generalization, it ensures object diversity across species, scales, and object counts, and provides purified camouflage-only annotations by filtering ambiguous labels to focus models on true camouflage patterns.

\subsection{Data Statistics}
The final DeepCamo dataset contains 2,493 underwater camouflaged images covering 16 marine species. It spans diverse lighting conditions and multi-object scenarios (see Fig.~\ref{first} for examples) and is split into DeepCamo-train (1,931 images) and DeepCamo-test (562 images) at an 8:2 ratio. Furthermore, we introduce DeepCamo-full, a benchmark subset of 1,907 images designed for fair and rigorous SOTA evaluation, which excludes all images overlapping with the COD10K training set to prevent data leakage. We hope DeepCamo will serve as a challenging and practical benchmark to advance research in UCOD.
\subsection{Benchmark Study}
We benchmarked 12 representative CNN-based and Transformer-based COD methods on our DeepCamo-full dataset, using their publicly available pretrained weights to ensure an objective assessment of generalization. As detailed in Table~\ref{tab:Benchmark_Study}, all methods struggled significantly. Key localization metrics ($S_{\alpha }$, $F_{\beta }$, $E_{\phi }$) dropped by over 15\%, while high MAE values indicated poor detail preservation. Even the top-performing model, SAM2-UNet, saw its MAE more than double (from 0.021 on COD10K to 0.045) and its $S_{\alpha }$ fall by 15.8\%. These findings reveal the stark generalization gap of current COD methods in underwater environments, underscoring the challenge and significance of our work.
\begin{table}
\centering
\caption{Quantitative comparison with SOTA on DeepCamo-full (best in bold).}
\begin{tabular*}{\textwidth}{@{\extracolsep{\fill}} c | l | c | c c c c}
\toprule
Category & Model & Pub./Year & $\mathcal{S}_\alpha \uparrow$ & $E_{\phi } \uparrow$ & $F^{w}_{\beta} \uparrow$ & MAE $\downarrow$ \\
\midrule
\multirow{6}{*}{CNN}
  & SINet~\cite{fan2020camouflaged} & CVPR/20 & 0.665 & 0.736 & 0.417 & 0.051 \\
  & SINet-V2~\cite{fan2021SINet-v2} & TPAMI/21 & 0.623 & 0.657 & 0.323 & 0.098 \\
  & PFNet~\cite{mei2021PFnet} & CVPR/21 & 0.674 & 0.756 & 0.428 & 0.049 \\
  & PreyNet~\cite{2022preynet} & ACM MM/22 & 0.677 & 0.738 & 0.444 & 0.058 \\
  & BCNet~\cite{xiao2023HFEM} & Neural Comput/23 & 0.683 & 0.744 & 0.460 & 0.067 \\
  & FEDER~\cite{2023FEDER} & CVPR/23 & 0.693 & 0.772 & 0.474 & 0.049 \\
\midrule
\multirow{6}{*}{Transformer}
  & PUENet~\cite{2023PUENet} & TIP/23 & 0.728 & \textbf{0.802} & 0.551 & 0.047 \\
  & FSNet~\cite{2023fsnet} & TIP/23 & 0.719 & 0.784 & 0.522 & 0.055 \\
  & HitNet~\cite{hu2023hitnet} & AAAI/23 & 0.712 & 0.765 & 0.527 & 0.061 \\
  & Dual-SAM~\cite{2024DualSAM} & CVPR/24 & 0.692 & 0.732 & 0.484 & 0.058 \\
  & MAMIFNet~\cite{2025MAMIFNet} & Inf. Fusion/24 & 0.720 & 0.784 & 0.530 & 0.054 \\
  & SAM2-UNet~\cite{xiong2024sam2UNet} & arXiv/24 & \textbf{0.741} & 0.801 & \textbf{0.557} & \textbf{0.045} \\
\bottomrule
\end{tabular*}
\label{tab:Benchmark_Study}
\end{table}
\section{Method}
\begin{figure}[t]
\includegraphics[width=\textwidth]{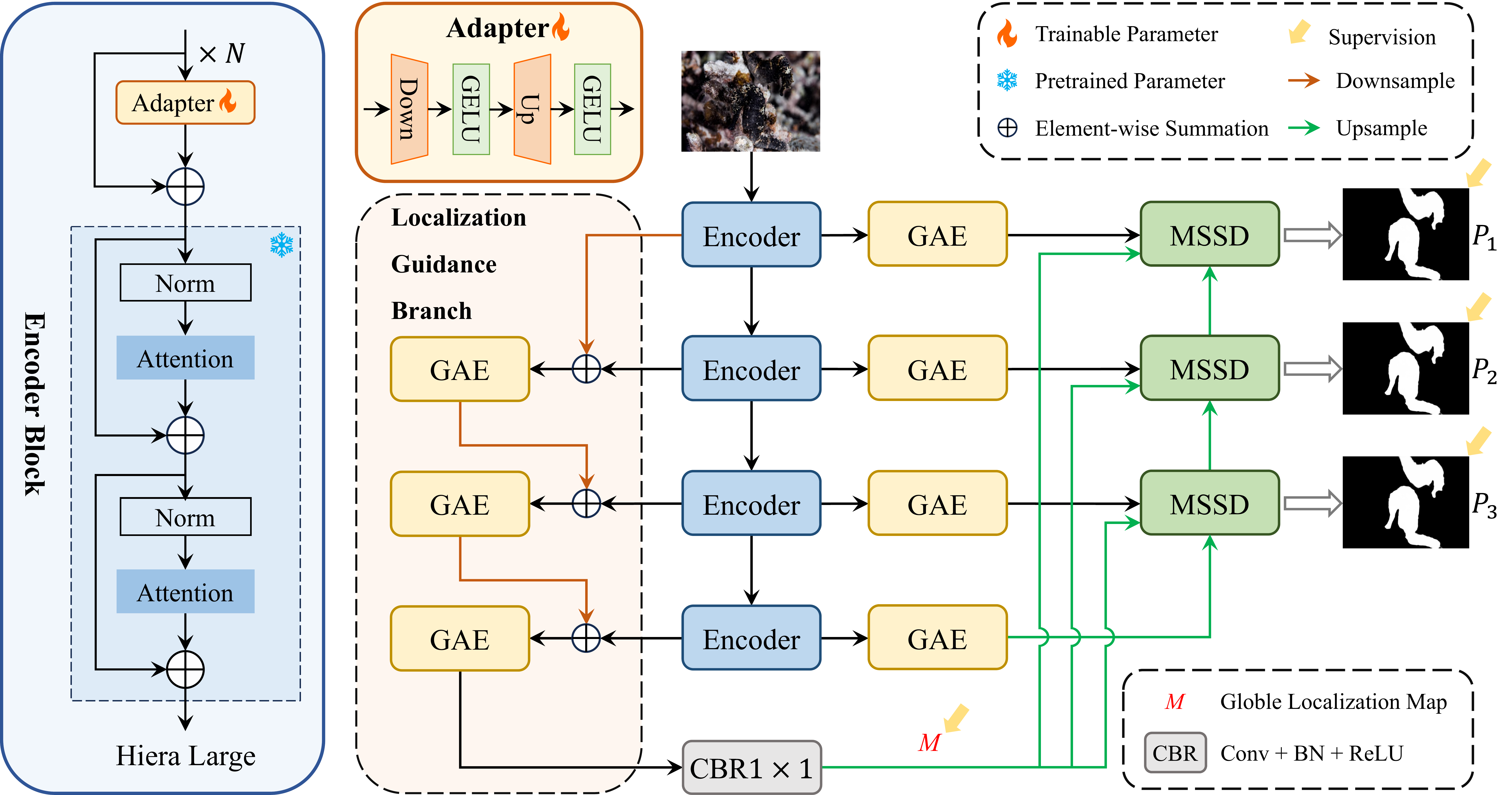}
\caption{The overall architecture of SLENet is composed of four key components: a SAM encoder with an Adapter, a Gamma-Asymmetric Enhancement (GAE) module, a Localization Guidance Branch (LGB), and a Multi-Scale Supervised Decoder (MSSD).} \label{max}
\end{figure}

UCOD is hindered by the loss of fine-grained details during multi-scale feature extraction and by inaccurate localization, which impairs the detection of small or multiple objects. To address these challenges, we propose SLENet (Fig.~\ref{max}), consisting of four key modules namely the SAM2 encoder with Adapters, the Gamma-Asymmetric Enhancement (GAE) module, the Localization Guidance Branch (LGB), and the Multi-Scale Supervised Decoder (MSSD). The following sections describe the design of each module and the overall loss function.
\subsection{SAM2 Encoder with Adapter}
To mitigate SAM2's tendency to produce category-irrelevant segmentations without manual prompts, we employ lightweight Adapters for parameter-efficient fine-tuning, injecting domain-specific knowledge for underwater camouflage. The backbone of SLENet adopts SAM2's Hiera-L encoder. While Hiera-L's parameters are frozen, we insert a trainable Adapter before each block to enable domain-specific tuning while capturing long-range context. Following established practices~\cite{houlsby2019adapter1,qiu2023lAdapter2}, each Adapter consists of a linear layer for downsampling, a GeLU activation, another linear layer for upsampling, and another GeLU activation. This Adapter-equipped encoder then extracts four levels of multi-scale features from an input image.
\subsection{Gamma-Asymmetric Enhancement Module}
To enhance multi-scale feature perception while preserving critical details, we propose the GAE Module. For each input feature $X_{i}, i \in\{1,2,3,4\}$, GAE employs four sequential branches. Each branch contains a $1\times1$ channel-reduction convolution, asymmetric convolutions to reduce parameter redundancy and enhance directional sensitivity, max-pooling layers, and a dilated convolution with a dilation rate of 2 to capture long-range dependencies. The output of each branch after channel compression is denoted as $X_i^r,\ r\in\left\{1,2,3,4\right\}$, where r is the branch index. For progressive enhancement, each subsequent branch contains fewer subsampling layers and receives the concatenated output of the previous branch as part of its input. This design expands the receptive field while integrating high-resolution features from earlier stages, thus enriching context and recovering fine details. The complete formulation for each branch is as follows:
\begin{align}
D_i^1 &=C_{dil}({AMP}_{\times3}(X_i^1)),\\
C_i^r &=C_{asy}(Cat(X_i^r,\ Up(D_i^{r-1},\ X_i^r))),\ r=2,3,4,\\
D_i^r &=C_{dil}({AMP}_{\times(4-r)}(C_i^r)),\ r=2,3,4,
\end{align}
where $C_{dil}$ and $C_{asy}$ denote dilated convolution and asymmetric convolution, respectively. ${AMP}_{\times N}$ denotes $N$ pairs of asymmetric convolution and max-pooling layers. $Up(D_i^{r-1},\ X_i^r)$ refers to upsampling $D_i^{r-1}$ to the spatial size of $X_i^r$, and $Cat$ denotes concatenation.
\begin{figure}[t]
\centering
\includegraphics[width=10cm]{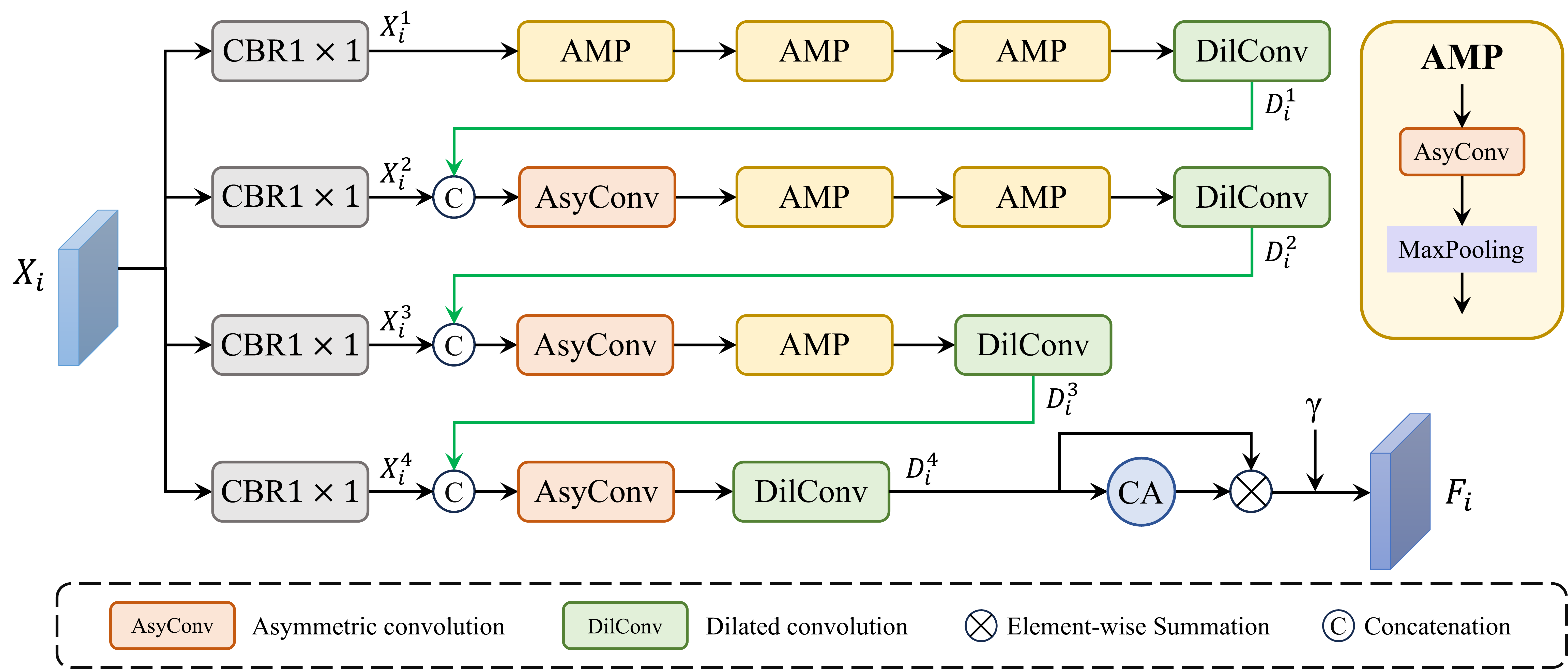}
\caption{The architecture of Gamma-Asymmetric Enhancement (GAE) module.} \label{GAE}
\end{figure}
The feature $D_i^4$ is processed by a channel attention mechanism to enhance discriminative semantic channels. Then, a learnable scaling factor $\gamma$ is introduced to adaptively adjust weights based on contextual information, enhancing both performance and stability. The final enhanced feature is thus obtained as:
\begin{equation}
F_i=\gamma(D_i^4\otimes CA(D_i^4)),
\end{equation}
where $CA$ is channel attention and $\otimes$ denotes element-wise multiplication.
\subsection{Localization Guidance Branch}
The Localization Guidance Branch (LGB) fuses cross-scale features in a bottom-up manner to refine the model's localization capabilities. Initially, the multi-scale features extracted from the backbone are compressed to a unified channel dimension via $1{\times}1$ convolutions and are denoted as $X_i^l$. Starting from the shallowest feature, each is fused with the adjacent deeper feature, then enhanced by the GAE module for improved contextual awareness. This enhanced output then serves as the input for the next fusion, as described by:
\begin{align}
F_2^l &=GAE(Down(X_1^l)\oplus X_2^l),\\
F_i^l &=GAE(Down(F_{i-1}^l)\oplus X_i^l),\ i=3,\ 4,
\end{align}
where $Down$ is downsampling, and $\oplus$ denotes element-wise summation. This iterative fusion process enables the LGB to effectively aggregate rich multi-scale semantic representations. The final feature $F_4^l$ is then processed by a $1\times1$ convolution to produce a coarse localization map $M$ with lower spatial resolution but stronger semantic representation:
\begin{align}
M = CBR_{1\times1}(F_4^l),
\end{align}
where ${CBR}_{N\times N}$ signifies a sequence of $N\times N$ convolution, batch normalization and ReLU.
\subsection{Multi-Scale Supervised Decoder}
\begin{figure}
\centering
\includegraphics[width=10cm]{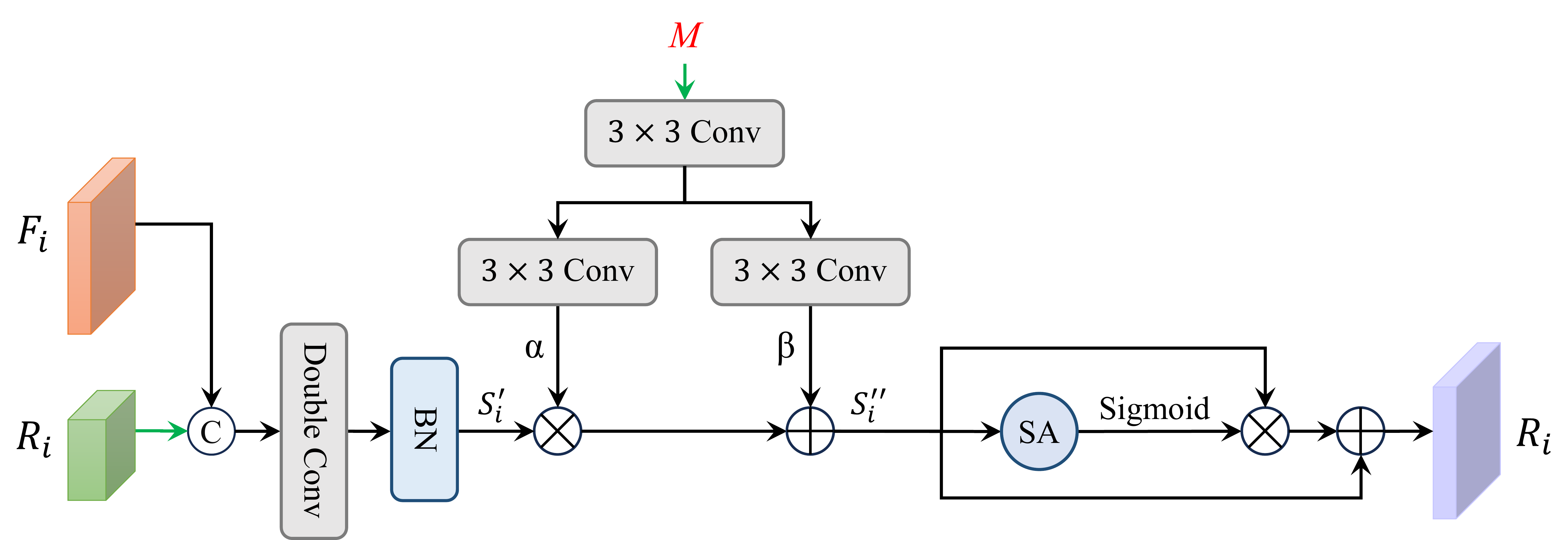}
\caption{The architecture of Multi-Scale Supervised Decoder (MSSD).} \label{MSSD}
\vspace{-1em}
\end{figure}

To enhance spatial awareness and focusing capabilities, the MSSD (Fig.~\ref{MSSD}) performs a top-down, iterative cross-scale fusion. It takes the GAE-enhanced features $F_i$ and the previous stage's output $R_{i+1}$ as input, and leverages the global localization map $M$ to achieve focus. Specifically, higher-level features are upsampled and concatenated with the current layer features $F_i$ along the channel dimension, then further fused through two consecutive $3\times3$ convolutional blocks to produce $S_i'$. This process is described as:
\begin{align}
S_i&=Cat\left(F_i,Up\left(R_{i+1},F_i\right)\right),\ i=1,2,\\
S_3&=Cat\left(F_3,Up\left(F_4,F_3\right)\right),\\
S_i^\prime&={CBR}_{3\times3}({CBR}_{3\times3}(S_i)),\ i=1,2,3.
\end{align}

\vspace{-2em}

\noindent The localization map $M$ serves as guidance for the feature modulation. It is first upsampled to match the feature resolution of $S_i'$ and then processed through two parallel $3\times 3$ convolutional blocks to generate a scale factorr $\alpha$ and shift factor $\beta$. These affine parameters are integrated into the normalization of the fused feature $S_i'$~\cite{2019semantic}, enabling the following modulation:
\begin{align}
S_i^{\prime\prime}&=BN(S_i^\prime)\otimes\alpha(M)\oplus\beta(M),
\end{align}
where $BN$ denotes batch normalization. A spatial attention block is then introduced to further enhance salient regions. The modulated features are combined with the weighted features via residual connection to produce the output $R_i$. The final segmentation result $P_i$ is obtained by applying a $1\times1$ convolution to $R_i$. This process is formulated as:
\begin{align}
R_i=(Sigmoid(SA(S_i^{\prime\prime}))\otimes S_i^{\prime\prime})\oplus S_i^{\prime\prime},\quad P_i={Conv}_{1\times1}(R_i),
\end{align}
where $SA$ denotes spatial attention, and $Conv_{1\times1}$ represents a $1\times1$ convolutional layer.
\subsection{Loss Function}
To mitigate the model's bias towards the background in class-imbalanced segmentation, we employ a weighted binary cross-entropy (wBCE) loss, with weights adaptively assigned based on pixel contrast. To further enhance global structure modeling, we incorporate a similarly weighted IoU loss, yielding the combined loss function $\mathcal{L}=\mathcal{L}_{BCE}^w+\mathcal{L}_{IOU}^w$.

The map $M$ is also supervised by an independent binary cross-entropy (BCE) loss. While this map provides crucial global guidance early in training, a persistently high loss weight can cause an over-reliance on its coarse representation, impeding the learning of fine-grained details. To mitigate this, we employ a linear decay strategy to dynamically adjust the weight of the $M$ branch. The specific weight and the final combined loss are defined as follows:
\begin{align}
\omega_m&=max(\mu(1-\frac{epoch}{epochs}),\ 0.1),\\
\mathcal{L}_{total}&=\omega_m\mathcal{L}_{BCE}(G,M)+\frac{1-\omega_m}{3}\sum_{i=1}^{3}{\mathcal{L}(G,\ P_i)},
\end{align}
the weight $\omega_m$ decays linearly from an initial value $\mu$ to 0.1 as the current epoch ($epoch$) progresses towards the total epochs ($epochs$). G denotes ground truth.

\section{Experiments}
\subsection{Experimental Setup}
Our model is implemented in PyTorch and trained on a single RTX 4090 GPU. We employ the AdamW optimizer with an initial learning rate of 5e-4, which decays following a cosine schedule. The backbone is the Hiera-L encoder from SAM2. All images are resized to 352×352 and processed with a batch size of 16 for 100 epochs. The weight parameter $\mu$ is set to 0.6. To evaluate our method, we conduct experiments on our proposed DeepCamo dataset and three public benchmarks, including CAMO~\cite{le2019CAMO}, CHAMELEON~\cite{2018Chameleon}, and COD10K~\cite{fan2020camouflaged}. Performance is assessed using five metrics: S-measure ($S_{\alpha }$), weighted F-measure ($F_\beta ^w$), mean E-measure ($E_{\phi }$), and mean absolute error (MAE).

\subsection{Comparison with SOTA Methods}
Our method is compared with 10 state-of-the-art COD approaches, including SINet~\cite{fan2020camouflaged}, SINet-V2~\cite{fan2021SINet-v2}, PFNet~\cite{mei2021PFnet}, BCNet~\cite{xiao2023HFEM}, FSNet~\cite{2023fsnet}, HitNet~\cite{hu2023hitnet}, PUENet~\cite{2023PUENet}, Dual-SAM~\cite{2024DualSAM}, MAMIFNet~\cite{2025MAMIFNet}, and SAM2-UNet~\cite{xiong2024sam2UNet}, on both the DeepCamo dataset and COD benchmark datasets.

\begin{table}[t]
\centering
\caption{Quantitative results on four benchmark datasets. `` $\uparrow/\downarrow$ '' indicates that larger or smaller is better. The best and second-best results are represented in bold and with an underline, respectively.}
\label{tab:sota}

\setlength{\tabcolsep}{4pt}
\begin{tabular}{l|c c c c|c c c c}
\toprule
\multirow{2}{*}{Models} & \multicolumn{4}{c|}{DeepCamo-Test} & \multicolumn{4}{c}{COD10K-Test} \\
 & $\mathcal{S}_{\alpha} \uparrow$ & $E_{\phi} \uparrow$ & $F^{w}_{\beta} \uparrow$ & MAE$\downarrow$ 
 & $\mathcal{S}_{\alpha} \uparrow$ & $E_{\phi} \uparrow$ & $F^{w}_{\beta} \uparrow$ & MAE$\downarrow$ \\
\midrule
SINet~\cite{fan2020camouflaged} & 0.745 & 0.803 & 0.501 & 0.057 & 0.771 & 0.806 & 0.551 & 0.051 \\
SINet-V2~\cite{fan2021SINet-v2} & 0.806 & 0.886 & 0.694 & 0.039 & 0.815 & 0.887 & 0.680 & 0.037 \\
PFNet~\cite{mei2021PFnet} & 0.805 & 0.880 & 0.690 & 0.041 & 0.800 & 0.868 & 0.660 & 0.040 \\
BCNet~\cite{xiao2023HFEM} & 0.815 & 0.897 & 0.716 & 0.035 & 0.837 & 0.894 & 0.704 & 0.033 \\
FSNet~\cite{2023fsnet} & 0.847 & 0.920 & 0.765 & 0.027 & 0.870 & \underline{0.938} & 0.810 & 0.023 \\
HitNet~\cite{hu2023hitnet} & 0.852 & 0.922 & \underline{0.778} & 0.030 & 0.868 & 0.932 & 0.798 & 0.024 \\
PUENet~\cite{2023PUENet} & 0.840 & 0.913 & 0.770 & 0.032 & 0.873 & \underline{0.938} & \underline{0.812} & 0.022 \\
Dual-SAM~\cite{2024DualSAM} & 0.819 & 0.873 & 0.733 & 0.034 & 0.832 & 0.893 & 0.754 & 0.027 \\
MAMIFNet~\cite{2025MAMIFNet} & 0.855 & \underline{0.925} & 0.774 & 0.028 & 0.869 & 0.933 & 0.785 & 0.023 \\
SAM2-UNet~\cite{xiong2024sam2UNet} & \underline{0.859} & 0.919 & 0.767 & \underline{0.025} & \underline{0.880} & 0.936 & 0.789 & \underline{0.021} \\
\textbf{SLENet(Ours)} & \textbf{0.869} & \textbf{0.930} & \textbf{0.800} & \textbf{0.022} & \textbf{0.883} & \textbf{0.945} & \textbf{0.820} & \textbf{0.019} \\
\bottomrule
\end{tabular}

\vspace{0.7em}

\begin{tabular}{l|c c c c|c c c c}
\toprule
\multirow{2}{*}{Models} & \multicolumn{4}{c|}{CAMO-Test} & \multicolumn{4}{c}{CHAMELEON} \\
 & $\mathcal{S}_{\alpha} \uparrow$ & $E_{\phi} \uparrow$ & $F^{w}_{\beta} \uparrow$ & MAE$\downarrow$
 & $\mathcal{S}_{\alpha} \uparrow$ & $E_{\phi} \uparrow$ & $F^{w}_{\beta} \uparrow$ & MAE$\downarrow$ \\
\midrule
SINet~\cite{fan2020camouflaged} & 0.751 & 0.771 & 0.606 & 0.100 & 0.869 & 0.891 & 0.740 & 0.044 \\
SINet-V2~\cite{fan2021SINet-v2} & 0.820 & 0.882 & 0.743 & 0.070 & 0.888 & 0.942 & 0.816 & 0.030 \\
PFNet~\cite{mei2021PFnet} & 0.782 & 0.852 & 0.695 & 0.085 & 0.882 & 0.942 & 0.810 & 0.033 \\
BCNet~\cite{xiao2023HFEM} & 0.830 & 0.886 & 0.761 & 0.068 & 0.901 & 0.944 & 0.839 & 0.029 \\
FSNet~\cite{2023fsnet} & 0.880 & \underline{0.933} & \textbf{0.861} & \underline{0.041} & 0.905 & 0.963 & 0.868 & 0.022 \\
HitNet~\cite{hu2023hitnet} & 0.844 & 0.903 & 0.801 & 0.057 & \underline{0.922} & \underline{0.970} & \textbf{0.903} & \underline{0.018} \\
PUENet~\cite{2023PUENet} & 0.877 & 0.930 & \underline{0.860} & 0.045 & 0.910 & 0.957 & 0.869 & 0.022 \\
Dual-SAM~\cite{2024DualSAM} & 0.825 & 0.885 & 0.787 & 0.058 & 0.807 & 0.877 & 0.731 & 0.048 \\
MAMIFNet~\cite{2025MAMIFNet} & 0.872 & 0.929 & 0.834 & 0.045 & 0.914 & 0.960 & 0.875 & 0.021 \\
SAM2-UNet~\cite{xiong2024sam2UNet} & \underline{0.884} & 0.932 & \textbf{0.861} & 0.042 & 0.914 & 0.961 & 0.863 & 0.022 \\
\textbf{SLENet(Ours)} & \textbf{0.887} & \textbf{0.937} & \underline{0.860} & \textbf{0.039} & \textbf{0.923} & \textbf{0.973} & \underline{0.896} & \textbf{0.017} \\
\bottomrule
\end{tabular}
\end{table}
\begin{figure}[!t]
\centering
\includegraphics[width=\textwidth]{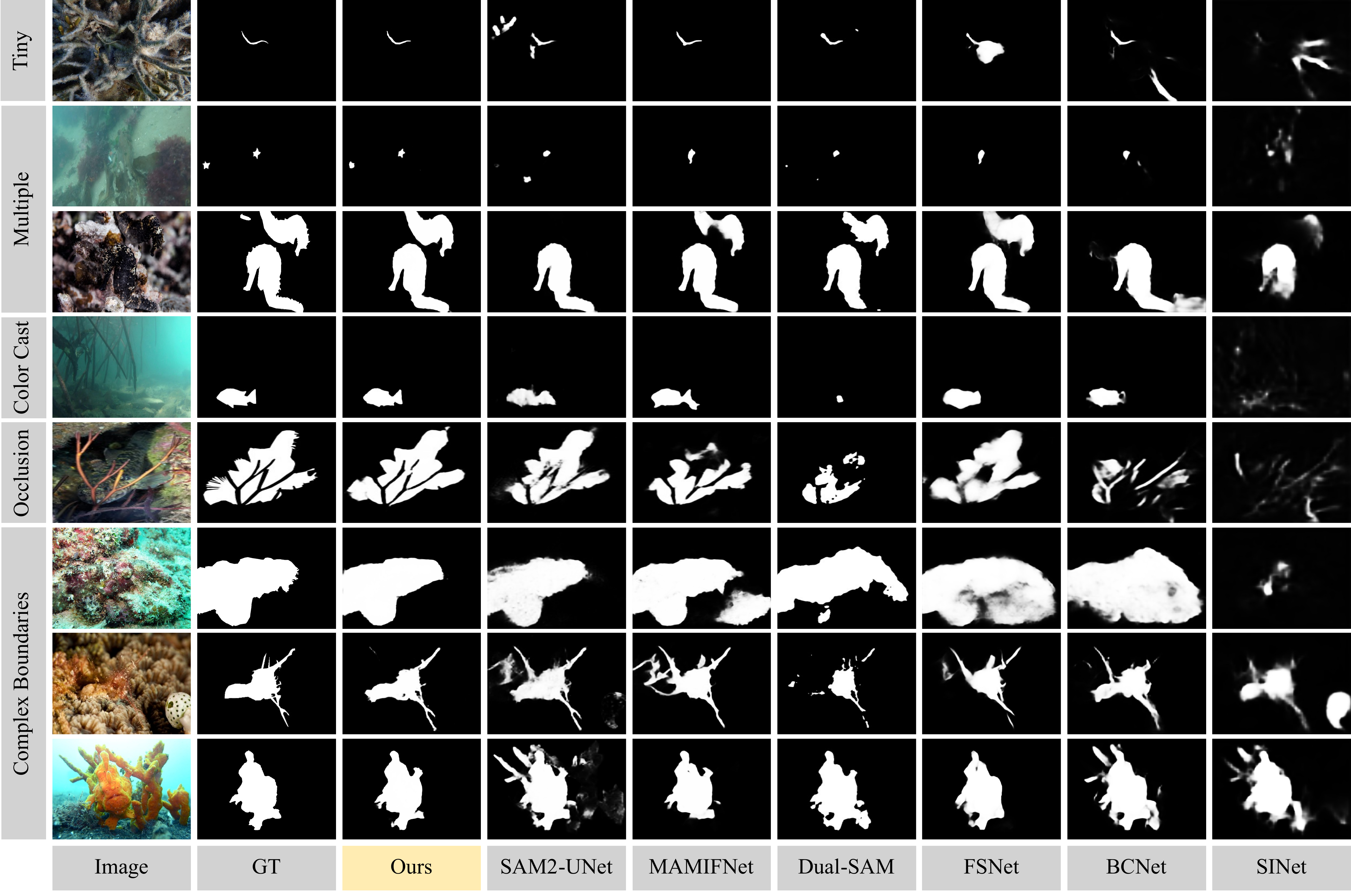}
\caption{Qualitative comparisons of the SLENet with several SOTA methods.} \label{Visualization}
\end{figure}
% \vspace{-1em}
\begin{table}[!t]
\centering
\caption{Ablation studies on the DeepCamo dataset. ``G'' ,``L'', and ``M'' denote GAE, LGB, and MSSD, respectively.}
\label{tab:ablation}

% Left table: Module ablation
\begin{minipage}[t]{0.45\linewidth}
\centering
\captionsetup{justification=centering}
\caption*{(a) Ablation results of each module.}
\setlength{\tabcolsep}{4pt}
\begin{tabular}{c c c | c c c c}
\toprule
G & L & M & $\mathcal{S}_\alpha \uparrow$ & $E_\phi \uparrow$ & $F^{w}_{\beta} \uparrow$ & MAE$\downarrow$ \\
\midrule
     &      &      & 0.858 & 0.918 & 0.764 & 0.026 \\
\checkmark &      &      & 0.864 & 0.928 & 0.784 & 0.023 \\
     & \checkmark & \checkmark & 0.864 & 0.920 & 0.791 & 0.024 \\
\checkmark & \checkmark & \checkmark & \textbf{0.869} & \textbf{0.930} & \textbf{0.800} & \textbf{0.022} \\
\bottomrule
\end{tabular}
\end{minipage}
\hfill
% Right table: Different μ values
\begin{minipage}[t]{0.45\linewidth}
\centering
\caption*{(b) Results with different $\mu$.}
\setlength{\tabcolsep}{4pt}
\begin{tabular}{c@{\hspace{8pt}} | c c c c}
\toprule
$\mu$ & $\mathcal{S}_{\alpha} \uparrow$ & $E_\phi \uparrow$ & $F^{w}_{\beta} \uparrow$ & MAE$\downarrow$ \\
\midrule
0.2 & 0.868 & 0.929 & 0.796 & 0.023 \\
0.4 & 0.866 & 0.928 & 0.795 & 0.023 \\
0.6 & \textbf{0.869} & \textbf{0.930} & \textbf{0.800} & \textbf{0.022} \\
0.8 & 0.865 & 0.924 & 0.791 & 0.022 \\
\bottomrule
\end{tabular}
\end{minipage}
\end{table}
\subsubsection{Quantitative comparison.} Table~\ref{tab:sota} summarizes the quantitative comparison of SLENet against SOTA methods on our DeepCamo dataset and three public benchmarks. SLENet consistently outperforms all competing CNN-based and Transformer-based methods across all evaluation metrics. This superiority is particularly evident on our challenging DeepCamo dataset, which features highly camouflaged objects and strong background interference. Here, SLENet surpasses the runner-up, SAM2-UNet, improving $S_\alpha$ and $F_\beta^w$ by 1.2\% and 4.3\% respectively, while decreasing the MAE by 12\%. Notably, achieving stable performance gains on a highly challenging dataset demonstrates the robustness of our model. Moreover, this strong performance extends to established benchmarks like COD10K, where SLENet achieves a 9.5\% reduction in MAE compared to SAM2-UNet. These consistent gains across diverse datasets underscore the effectiveness and strong generalization capabilities of SLENet.
\subsubsection{Qualitative comparison.} Fig.~\ref{Visualization} showcases the qualitative performance of SLENet on representative samples from various test sets. These samples cover challenging scenarios, including small or multiple objects, occlusion, and poor visibility due to distortion or turbidity. Across these complex scenes, SLENet accurately identifies object regions and delineates their boundaries with high precision. In contrast, many SOTA methods struggle with such cases, often yielding missed detections or blurred edges. By incorporating the GAE and LGB modules, our model effectively reconstructs fine-grained details and improves localization accuracy, yielding predictions that closely match the ground truth. This visual evidence clearly demonstrates the superiority of SLENet over existing approaches.
\subsection{Ablation Study}
\subsubsection{Impact of key components.} Ablation studies on DeepCamo (Table~\ref{tab:ablation}a) assess the impact of key components (GAE, LGB, and MSSD) by removing them from the full model. 
\begin{itemize}
    \item \textbf{Effectiveness of GAE.} A baseline model using only plain convolutions performs the worst. Adding GAE significantly enhances context extraction and detail modeling, improving $F_\beta^w$ by 2.6\% and reducing MAE by 11.5\%. This demonstrates its ability to capture fine boundary and texture details.
    
    \item \textbf{Effectiveness of LGB \& MSSD.} LGB offers global localization guidance. Combined with MSSD for multi-scale fusion, it improves localization and structural consistency, yielding a 3.5\% $F_\beta^w$ gain over the baseline.

    \item \textbf{Effectiveness of full model.} The full model achieves top performance across all metrics, confirming their synergistic contributions to precise localization and context-aware representation in complex environments.
\end{itemize}

\subsubsection{Impact of $\mu$.} We analyze the impact of $\mu$, the initial loss weight for the LGB's global localization map $M$ (Table~\ref{tab:ablation}b). When $\mu=0$,  the $M$ branch weight is fixed at a low 0.1, providing minimal guidance and leading to a significant performance drop. This result is omitted from the table to better illustrate the key trend. As $\mu$ increases, enhanced global guidance improves object localization, with optimal performance achieved at $\mu=0.6$. However, excessively high $\mu$ values cause over-reliance on global supervision, thereby hindering fine-grained feature learning and degrading performance.

\section{Conclusion}
In this work, we present a comprehensive study of the challenging task of underwater camouflaged object detection. To facilitate research in this underexplored domain, we introduce DeepCamo, a benchmark dataset tailored for UCOD, which establishes a foundational benchmark. We also propose SLENet, a novel framework tailored to the unique challenges of underwater camouflage. SLENet introduces the GAE module to enhance features across all scales, significantly improving detail and texture modeling. Building on this, the LGB and MSSD modules work in concert to provide global localization modulation and cross-scale feature fusion. Extensive experiments on DeepCamo and three general COD datasets validate that SLENet achieves SOTA performance, demonstrating excellent robustness and generalization. We believe this work provides a solid foundation for future UCOD research, with promising directions including underwater image enhancement and cross-domain adaptation.

\bibliographystyle{splncs04}
\bibliography{original_materials/reference}

\end{document}